%% file: main.tex
\documentclass{sigkddExp}

\usepackage{placeins}

\newcommand{\num}{five}

\newcolumntype{P}[1]{>{\centering\arraybackslash}p{#1}}

\definecolor{users}{RGB}{228,26,28}
\definecolor{transforms}{RGB}{152,78,163}
\definecolor{uses}{RGB}{220,220,20}

\definecolor{properties-highlight}{RGB}{219,255,255}
\definecolor{properties-text}{RGB}{38,129,139}

\definecolor{users-highlight}{RGB}{255,217,214}
\definecolor{users-text}{RGB}{166,66,39}

\definecolor{transforms-highlight}{RGB}{233,232,255}
\definecolor{transforms-text}{RGB}{44,35,95}

\newcommand{\property}[1]{\sethlcolor{properties-highlight}\textcolor{properties-text}{\hl{#1}}}

\newcommand{\user}[1]{\sethlcolor{users-highlight}\textcolor{users-text}{\hl{#1}}}

\newcommand{\transform}[1]{\sethlcolor{transforms-highlight}\textcolor{transforms-text}{\hl{#1}}}

\begin{document}
%
% --- Author Metadata here ---
% -- Can be completely blank or contain 'commented' information like this...
%\conferenceinfo{WOODSTOCK}{'97 El Paso, Texas USA} % If you happen to know the conference location etc.
%\CopyrightYear{2001} % Allows a non-default  copyright year  to be 'entered' - IF NEED BE.
%\crdata{0-12345-67-8/90/01}  % Allows non-default copyright data to be 'entered' - IF NEED BE.
% --- End of author Metadata ---

\title{The Need for Interpretable Features: Motivation and Taxonomy}
%\subtitle{[Extended Abstract]
% You need the command \numberofauthors to handle the "boxing"
% and alignment of the authors under the title, and to add
% a section for authors number 4 through n.
%
% Up to the first three authors are aligned under the title;
% use the \alignauthor commands below to handle those names
% and affiliations. Add names, affiliations, addresses for
% additional authors as the argument to \additionalauthors;
% these will be set for you without further effort on your
% part as the last section in the body of your article BEFORE
% References or any Appendices.

\numberofauthors{1}
%
% You can go ahead and credit authors number 4+ here;
% their names will appear in a section called
% "Additional Authors" just before the Appendices
% (if there are any) or Bibliography (if there
% aren't)

% Put no more than the first THREE authors in the \author command
%%You are free to format the authors in alternate ways if you have more 
%%than three authors.

\author{
%% e-mail address with \email.
\alignauthor Alexandra Zytek\textsuperscript{*}, Ignacio Arnaldo\textsuperscript{\dag}, Dongyu Liu\textsuperscript{*}, Laure Berti-{É}quille\textsuperscript{\ddag}, Kalyan Veeramachaneni\textsuperscript{*} \\
       \affaddr{\textsuperscript{*}MIT, Cambridge MA, USA}\\
       \affaddr{\textsuperscript{\dag}Corelight} \\
       \affaddr{\textsuperscript{\ddag}IRD, ESPACE-DEV }\\
       \email{\{zyteka, dongyu, kalyanv\}@mit.edu}
      \email{nachoarnaldo@gmail.com}
       \email{laure.berti@ird.fr}
}

\maketitle
\begin{abstract}
Through extensive experience developing and explaining machine learning (ML) applications for real-world domains, we have learned that ML models are only as interpretable as their features. Even simple, highly interpretable model types such as regression models can be difficult or impossible to understand if they use uninterpretable features. Different users, especially those using ML models for decision-making in their domains, may require different levels and types of feature interpretability. Furthermore, based on our experiences, we claim that the term ``interpretable feature'' is not specific nor detailed enough to capture the full extent to which features impact the usefulness of ML explanations. In this paper, we motivate and discuss three key lessons: 1) more attention should be given to what we refer to as the \textit{interpretable feature space}, or the state of features that are useful to domain experts taking real-world actions, 2) a formal taxonomy is needed of the feature properties that may be required by these domain experts (we propose a partial taxonomy in this paper), and 3) transforms that take data from the model-ready state to an interpretable form are just as essential as traditional ML transforms that prepare features for the model.
\end{abstract}

\input{textv2/introduction}
\input{textv2/literature_review}

\input{textv2/taxonomy}

\input{textv2/conclusion}
%ACKNOWLEDGEMENTS are optional
\section{Acknowledgements}
The Forest Covertype dataset is copyrighted 1998 by Jock A. Blackard and Colorado State University.

\bibliographystyle{abbrv}
\bibliography{references_zot}  % sigproc.bib is the name of the Bibliography in this case
% ACM needs 'a single self-contained file'!
% SIGKDD: balancing columns messes up the footers: Sunita Sarawagi, Jan 2000.
% \balancecolumns
%\appendix
%Appendix A

% That's all folks!
\end{document}

%% file: textv2/introduction.tex
\section{Introduction} \label{sec:introduction}

\begin{figure*}
    \centering
    \includegraphics[width=1.0\linewidth]{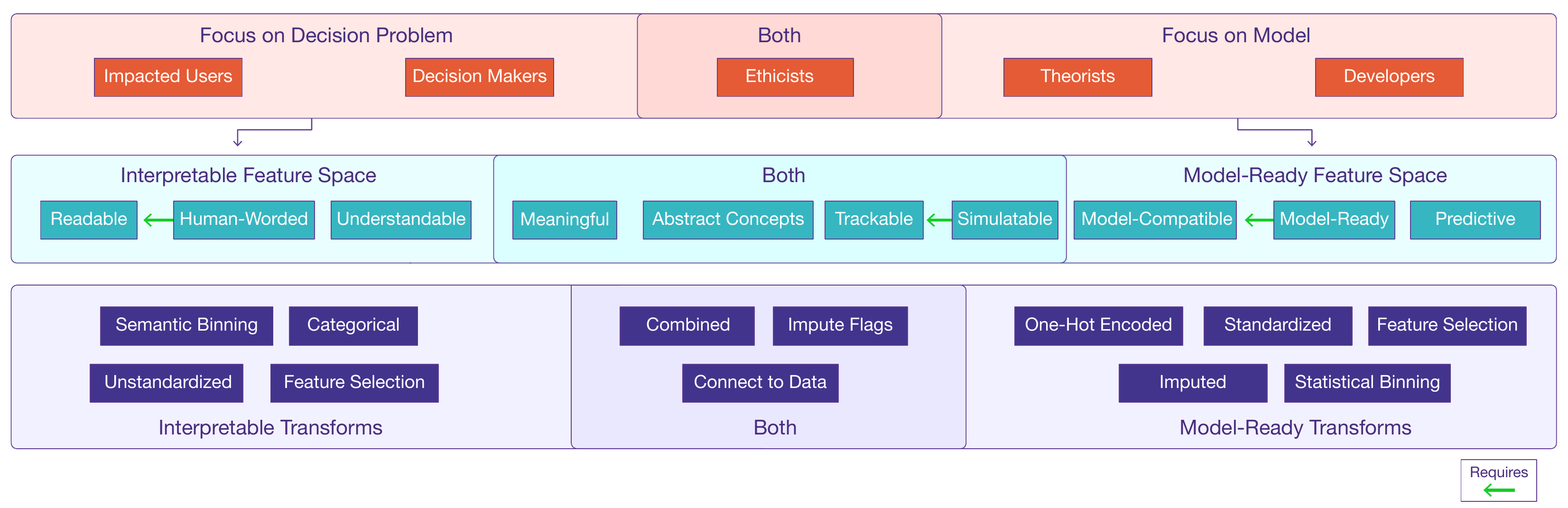}
    \caption{Summary of our feature taxonomy and its connection to other elements of the ML workflow. This table shows the different users (top layer), the feature properties required for their workflows (middle layer), and the transforms that yield the required feature properties (bottom layer). The green lines between properties indicate that, for a feature to have the property at the head-end of the arrow, it must also have the property at the tail-end of the arrow. An example of a path through this figure: \user{decision makers} usually require the interpretable feature space, which is characterized by features that are \property{meaningful}, \property{understandable}, \property{human-worded}, and/or \property{readable}. To achieve these characteristics, model explanations should rely on transform states such as \transform{unstandardized}, \transform{categorical}, and \transform{semantic binning}, and should avoid model-ready transformations such as \transform{unflagged imputations} or \transform{one-hot encoding}}.
    \label{fig:summary}
\end{figure*}

\input{Figures/application_table_2}

Much work has been done on developing models and algorithms to improve the interpretability and explainability of ML for tabular data.\footnote{There remains disagreement among the community as to the exact distinction between the terms \textit{interpretable} and \textit{explainable} ML. Our discussion in this paper refers to the ability for humans to understand how a model makes its decisions --- either through the use of naturally interpretable models or post-hoc explanations \cite{doshi-velez_towards_2017}. We will use both terms to capture this concept.} However, our experiences introducing ML solutions to a wide variety of real-world domains \cite{arnaldo_holy_2019} \cite{cheng_vbridge_2021} \cite{liu_mtv_2021} \cite{veeramachaneni_towards_2014} \cite{zytek_sibyl_2021}  has shown that models are only as interpretable as their features. ML explanations are presented using the language of the model features; whether through presenting feature importances, meaningful example inputs, decision boundary visualizations, or other methods. While the literature has acknowledged the importance of interpretable features (IFs) and suggested a wide variety of methods to generate them (see Section \ref{sec:literature_review}), we have found a lack of a nuanced and context-aware analysis of what it means for a feature to be ``interpretable'' for a given purpose. In particular, we find that an ``interpretable feature'' is often defined simply as one that was human-generated or worded using human-readable language. However, our extensive experiences in \num{} diverse domains have suggested that this definition is incomplete, and for features to be  useful as part of ML explanations and other augmenting information, they may require a variety of properties. 

In this paper, we aim to better define the concept of feature interpretability for domains using tabular data and time series data that is engineered into tabular features. Our experiences, as described below, mostly revolve around this data modality. Other modalities, such as image and text are unique enough to warrant a separate discussion on the interpretability of features.

%Additionally, we have found a lack of a formal taxonomy that can be used to identify the properties that may increase and decrease the interpretability of features in specific contexts. 

In part, the lack of formal investigation into the properties that make features interpretable stems from the tendency for literature to focus on only a subset of possible ML stakeholders --- mostly ML developers and theorists, who have expertise in ML and data science and are interested in designing, improving, and validating models. For these groups, the goal is to engineer (or discover, such as through deep learning techniques) features that maximize model performance, often without consideration for whether the designed features are meaningful to domain experts. They are more interested in understanding how features interact with and impact the ML model than their connections to the real-world domain.

Preece et. al. (2018) introduced four major user types to ML and ML explanations \cite{preece_stakeholders_2018}; we modified and extended this list based on our experiences. We define these users by their goals related to ML explanations.

\begin{itemize}
    \item \user{Developers} \cite{preece_stakeholders_2018} build, train, test, and deploy ML solutions. They are most interested in understanding the feature space used by the model, and their primary goal is to maximize model performance.
    \item \user{Theorists} \cite{preece_stakeholders_2018} are interested in understanding and advancing ML theory. Unlike developers, they are more interested in improving ML algorithms and tools rather than focusing on specific applications. Their goal is to understand how models themselves work, and may therefore be more interested in white-box explanations that elucidate models' inner workings. They are also usually interested in understanding the feature space used by the model.
    \item \user{Ethicists} \cite{preece_stakeholders_2018} are interested in the fairness, accountability, and transparency of ML systems. This group may contain a diverse range of domain experts, including computer scientists, social scientists, and policy-makers. This group may be interested in features as used by the model, or in a more interpretable form, depending on their specific goal and expertise. While ethicists from the ML-expertise perspective get attention in the ML fairness literature, we found relatively less attention for ethicists from domain-expertise perspectives. 
    \item \user{Decision makers} use ML models and explanations to help them complete tasks and make decisions. In most domains, this group will not have ML expertise.
    \item \user{Impacted Users} are affected by ML models and their use, but do not directly interact with the models or their predictions --- except, possibly, to understand or argue against their impact on them. They rarely have ML expertise, and often will also not have domain expertise. 
\end{itemize}

Decision makers, impacted users, and some ethicists are especially likely to need explanations that use features that have a deeper level of interpretability. In this paper, we discuss our experiences working mostly with decision makers, as well as some domain-expert ethicists, and the feature interpretability properties that we have found necessary for these groups.

Figure \ref{fig:summary} summarizes the contributions of this paper, and highlights the difference between the needs of users focused on the model versus users focused on real-world decision-making. The figure includes the possible ML users, the feature properties introduced in this paper, and some relevant transforms that may offer these properties. The figure suggests a set of paths for interpretable feature engineering based on user needs.

We will now introduce the \num{} real-world domains in which we have experience developing ML solutions or explanation tools. In each domain, we worked with domain experts (who did not have ML experience) to introduce and explain one or more ML models. In all cases, we found that a major roadblock to model usefulness stemmed from confusion related to the features. We will continue providing examples of these roadblocks throughout the paper as we introduce our feature property taxonomy. These domains are summarized in Table \ref{tab:applications}.

% Domain example outline:
%   . scenario
%   . stakeholders
%   . prediction task
%   . data and features
%   . process
%   . challenge summary

\subsection{Child Welfare: Clarifying Features}
Our first example is in the domain of child welfare screening \cite{zytek_sibyl_2021}. Our team collaborated with a team of social workers and social scientists over the course of a year to introduce ML explanations alongside the existing ML solution. These users filled the role of \user{decision makers} making screening decisions, and \user{ethicists} ensuring that the model was working in a fair and holistic manner. The decision problem at hand was whether or not to screen-in a case of potential child abuse for further investigation. In our collaborating counties, these experts were provided with an ML-predicted 1 through 20 risk score, which they could use to aid their usual decision making process. The model is a lasso linear regression takes in over 400 hand-generated features based on data such as demographic information, criminal history, and child welfare referral history. 

We ran a series of formal user studies, where child welfare screeners were provided with feature contribution explanations alongside the predicted risk score to aid them in their screening decision-making. We found that most confusion and distrust in the model stemmed from the features themselves, rather than details of the model or explanation type. This was despite the features being interpretable by the usual definitions found in the literature (as elaborated on in Section \ref{sec:literature_review}) --- they were hand-selected by humans, and presented in readable, natural language. The main challenges related to the features being worded confusingly (for example, \texttt{role of child is sibling is FALSE}, or being seemingly unrelated to the prediction target according to users (such as features related to the family getting food aid). 

\subsection{Education: Abstracting Features}

In the domain of online education, we have worked on feature engineering and prediction tasks on data related to massive open online classes (MOOCs) \cite{veeramachaneni_towards_2014}. This research applies to a diverse set of users, including \user{ML developers} working on MOOC analysis systems, course developers looking to learn about and improve online course quality (\user{decision makers}), and instructors wanting to better understand their student cohort (\user{decision makers}). The prediction tasks related to MOOCs include classifying student engagement level and type and identifying course resource usage. The features used in these tasks include data such as problem set grades, hours of videos watched, or number of discussion posts.

We worked through the process of engineering MOOC features to better inform interested parties about the students and classes. Our experiences and the literature on this topic have shown that the most useful features take the form of abstract concepts that have meaning to the users (such as summarizing features relating to completing work and interacting with course materials as \texttt{participation}), as well as demonstrating the importance of being able to fully understand the source of these concepts.

\subsection{Cybersecurity: Tracking Features}

Our next example is in the domain of cybersecurity --- specifically, the detection of Domain Generation Algorithms (DGAs) \cite{arnaldo_holy_2019}, a well-documented command and control mechanism~\cite{plohmann_comprehensive_2016}. However, the lessons learned are applicable to a wide range of threat detection use cases.
DGAs are designed by attackers to generate dynamic domains that are then used as meeting points for compromised machines and remote attackers. The key idea is that the remote attacker and the compromised machine share the generation logic, and will therefore generate the same set of domains, each on its own side. The remote attacker then registers one of the many domains generated by the algorithm, while the malware in the compromised machines attempts to connect to each and every one of the generated domains until a communication channel is established with the remote attacker \cite{arnaldo_holy_2019}. In this case, our users were security analysts, \user{making decisions} about how to respond to a potential attack. The model was trained to detect and flag potential DGA attacks based on aggregated information (features) about the activity of internal IPs captured in DNS logs (such as the number of failed DNS requests, or the randomness of queried domains).

We found that the raw data --- in this case, the relevant sections of DNS logs --- were more useful to the users than the generated features, as the logs held more familiar and actionable information. For example, analysts need to investigate the specific domains queried by a suspected IP to determine the presence or absence of a DGA.
In this case, the challenge for explainability was how to select these relevant logs, to accurately track feature values back to the raw data they were generated from.

\subsection{Healthcare: Disaggregating Features}
In the domain of healthcare, we built ML prediction models using Electronic Health Record (EHR) data to improve the quality of clinical care \cite{cheng_vbridge_2021}. In particular, we closely worked with six clinicians (\user{decision makers}) with an average of 17.5 years of work experience. The task was predicting complications after cardiac surgery. The clinicians were interested in using ML models to predict the chance of a patient developing five types of complications (lung, cardiac, arrhythmia, infectious, and others) after surgery. To this end, we extracted various types of ML features from the medical records of target patients, mainly including three types of static features (demographics, surgery information, and diagnosis results) and three types of time-varying features (lab tests, routine vital signs, and surgery vital signs). We then built five individual binary classifiers with each predicting one of the five complication types.

We adopted Shapley Additive Explanations (SHAP) \cite{lundberg_unified_2017}, a game-theoretical method to calculate the feature contributions. However, simply providing SHAP explanations to clinicians is difficult. We found that, in particular, some features based on aggregation functions are not as interpretable as the original signal data.

%We identified two particular challenges having domain experts (i.e., clinicians) to understand the explanations.

%\begin{itemize}[noitemsep,nolistsep]
%\item {\textbf{Some ML features are not interpretable as-is by clinicians}. For example, we applied aggregation functions on time-varying features and obtain aggregate values (e.g., standard deviation or linear slope) within a period. Though intuitive to ML models, clinicians are struggle to understand what the feature values indicate and what the potential consequences are.}
%\end{itemize}

\subsection{Satellite Monitoring: Marking Real Features}
In the domain of satellite monitoring, we approached the challenge of visualizing the results of ML time-series anomaly detection solutions \cite{liu_mtv_2021}. While this domain used only time-series data, rather than tabular data, we learned lessons about tracking imputations that extend to the tabular use case. We developed a visualization tool that allowed domain experts to interact with ML predictions, and analyze and discuss the relevance and validity of predicted anomalies. We actively collaborated with members of our intended user group --- six domain experts with experience in anomaly detection \user{decision making}. The ML model we employed for the tool makes use of time-series sensor data to detect anomalous periods.

We developed the tool with active feedback from our six domain experts. We then ran two user studies to evaluate the tool, first with the six domain experts, and then with a group of 25 general end-users using stock price data. This process revealed feature-relevant considerations, such as the need to clarify which feature values were the result of an imputation process, and therefore not real values.

\subsection{Key Lessons}
Our experiences in the domains described above have led to three key lessons, which we elaborate on in this paper. \\

\textbf{Lesson 1: We Need an Interpretable Feature Space}
More focus is needed on what we refer to as the \textit{interpretable feature space}, or the information used by the model transformed to a human-usable set of features. Much attention in the literature has been given to collecting, selecting, and engineering features \cite{khalid_survey_2014} \cite{zheng_feature_2018}, usually with the goal of maximizing model test-set and real-world performance. The result is a heavy focus on the transformations that convert from the \textit{original feature space} to the \textit{model-ready feature space}. However, when an ML model needs to interface with human users, especially for active decision-making, the interpretable feature space may be needed. This feature space is engineered to closely represent reality and human cognition rather than ML and statistical concepts. More work is needed on how to define, generate, and evaluate this feature space --- this paper focuses on defining. \\

\textbf{Lesson 2: Features Require Certain Properties to Be Interpretable}
The properties required for features to be useful heavily depends on context and use case. As suggested in Lesson 1, if the goal is only model performance, important properties may include that features are in a model-compatible format and statistically correlated with the target variable. However, if the goal is provide useful explanations to decision-makers, features that understandably relate to real-world and abstract concepts may be more useful. To properly understand these properties, we propose a formal taxonomy of feature properties, elaborated on in Section \ref{sec:taxonomy}. \\

\textbf{Lesson 3: We Need to Transform the Other Way}
Feature properties are achieved through a combination of feature selection and engineering. Transforms that bring features to a model-ready state may increase or decrease their interpretability; in the latter case, model-ready transforms may need to be \textit{undone} for the purposes of explanations. In addition, further transforms for features presented in explanations may improve their interpretability. However, many ML libraries do not include functionality for transforms that improve interpretability or undo model-ready transforms. Section \ref{sec:transforms} provides a partial list of such transforms. 

%% file: Figures/application_table_2.tex
% \usepackage{multirow}
% \usepackage[normalem]{ulem}
\renewcommand{\arraystretch}{1.7}
\begin{table*}[ht]
\caption{Summary of example applications.}
\label{tab:applications}
\begin{tabular}{p{0.24\linewidth}p{0.33\linewidth}p{0.37\linewidth}}
\hline
\textbf{Domain}           & \textbf{Stakeholder Type}                                                                                                                                                                                            & \textbf{Description}                                                                                                                                                                                                                                                                                                                                                                                                                                                                    \\ \hline
Child Welfare Screening \cite{zytek_sibyl_2021}   & \textbf{Decision Makers:} \newline \hspace*{1ex} Social workers screening cases   \newline \textbf{Ethicists:}  \newline \hspace*{1ex} Social scientists reviewing the model                                                                                 & Collaborated with social scientists and social workers to develop an ML explanation toolkit. The toolkit provides detailed insights about feature contributions and distributions.                                                                                                                                             \\ 
Education \cite{veeramachaneni_towards_2014}                & \textbf{Decision Makers:} \newline \hspace*{1ex} Instructors conducting courses \newline \hspace*{1ex} Instructors designing courses \newline \textbf{Ethicists:} \newline \hspace*{1ex} Domain experts reviewing courses  \newline \textbf{ML Developers:} \newline \hspace*{1ex} Developers working on analysis systems      & A series of different experiments involving feature engineering and analyzing outcomes from MOOCs                                                                                                                                                                                                                       \\ Cybersecurity \cite{arnaldo_holy_2019}            & \textbf{Decision Makers:} \newline \hspace*{1ex} Domain experts monitoring for attacks                                                                                                                                                                              & Experience in developing, evaluating, and deploying machine learning for cybersecurity. Extended the detection capability of security operations centers with machine learning models trained to identify attacks in log data.                                                                                                                                                                                                                                                                                                                                                                                                                                                                                                                         \\
Electronic Health Records \cite{cheng_vbridge_2021} & \textbf{Decision Makers:} \newline \hspace*{1ex} Doctors conducting surgery                                                                                                                                                                          & Collaborated with doctors to develop a visualization tool to explain predictions on EHR data. The tool displays feature contributions, and shows the original raw signal data that features are generated from.                                                                                     \\
Satellite Monitoring \cite{liu_mtv_2021}     & \textbf{Decision Makers:} \newline \hspace*{1ex} Domain experts monitoring signals                                                                                                                                                              & Collaborated with satellite experts to develop a tool to visualize ML-predicted anomalies in satellite signal data.                                                                                                                                                                                                                                                                                                                                 \\
 \hline
\end{tabular}
\end{table*}

%% file: textv2/literature_review.tex
\section{Related Work} \label{sec:literature_review}

\input{Figures/literature_review_table}

\subsection{Formal Literature Review}
To make progress on understanding the existing perspectives on interpretable features in the literature, we conducted a methodological literature review. We reviewed papers from 19 top conferences and journals in the fields of machine learning, human-computer interaction, computer vision, and data visualization. For each of these venues, we searched for papers that included (in title or text body) the keywords ``machine learning'' or ``ML'', as well as the phrase ``interpretable feature(s)'' or any of its semantic equivalents (readable/understandable/explainable feature(s), interpretability of (the) feature(s)). We limited our search to papers that were published during or after 2015, as this can be considered the start of the newest wave of interest in explaining black-box ML \cite{molnar_interpretable_2020}. In total, our literature review included 106 papers.\footnote{For comparison, we ran a larger search using the same procedure for all papers published in the same years and venues described above that mentioned ``machine learning'' or ``ML'' and the word ``features'' in any context. This search revealed a total of 13,104 papers that talked about features and ML.}

For each paper, we noted the domain, data type, and goal of the paper, as well as the reason for mentioning interpretable features, which we then codified into 8 categories, seen in Table \ref{tab:literature_review}. We also took note of the papers that provided any kind of definition, either general or context-specific, for the concept of ``interpretable features''.

Our findings from our literature review can be summarized as follows. \\

\textbf{Attention is given to the importance of IFs.}  The need for or importance of IFs is well documented in the literature. 47 papers indicated the importance of IFs, either in general or to the methods proposed in the paper.  49 papers claimed to generate interpretable features through their proposed methods, though only 18 offered any kind of concrete definition of an interpretable feature. Additionally, five papers explicitly stated that their algorithm or model generated uninterpretable features. \\
    
\textbf{Little work on IFs addresses tabular data.} Of the 106 papers in our literature review, 47 mentioned interpretable features only briefly with no further elaboration as to how or why interpretable features were chosen or used. 80 focus on image, text, or time series data, and/or discuss interpretable features in the context of interpreting the latent features discovered by neural networks. 26 papers provided concrete insights on interpretable input features for tabular or arbitrary data types. \\

\textbf{There is a lack of concrete definitions for IFs.} Only 30 out of 106 papers offered at least a brief general or context-specific definition or quantification method for interpretable features. Of these, seven defined interpretable features as being those selected or engineered by humans, or similar to those selected by humans. Four defined interpretable features by contrasting them to the uninterpretable latent features used by neural networks or PCA. Four papers considered features to be more interpretable when related to abstract concepts. Only one paper \cite{yadav_explanation_2017} directly quantified the interpretability of features through a user study. \\

Our conclusion from our formal literature review is that although there is wide-spread agreement that interpretable features are important for various reasons\footnote{It is worth noting that two papers in our literature review did find that in contexts where privacy is important, uninterpretable features may be preferable - see the last row of Table \ref{tab:literature_review}.}, there is little work formalizing what makes a feature interpretable, and almost no work quantifying the interpretability of features. Most of the concrete work on interpretable features has focused on complex data domains such as image and text, and aim to add interpretability to deep neural networks. Given our experiences in the \num{} domains described in our introduction, we claim that a deeper, human-centered understanding of interpretable features for tabular features is necessary. 

We have also found that the term \textit{interpretable feature} does not have a clearly agreed upon meaning, but usually refers to features that are human-generated or can be understood by humans in the most surface-level way (i.e., humans can understand what a feature refers to). In this paper, we seek to go beyond the term \textit{interpretable} to introduce a complete, descriptive taxonomy of feature properties. 

\begin{figure*}
    \centering
    \includegraphics[width=\linewidth]{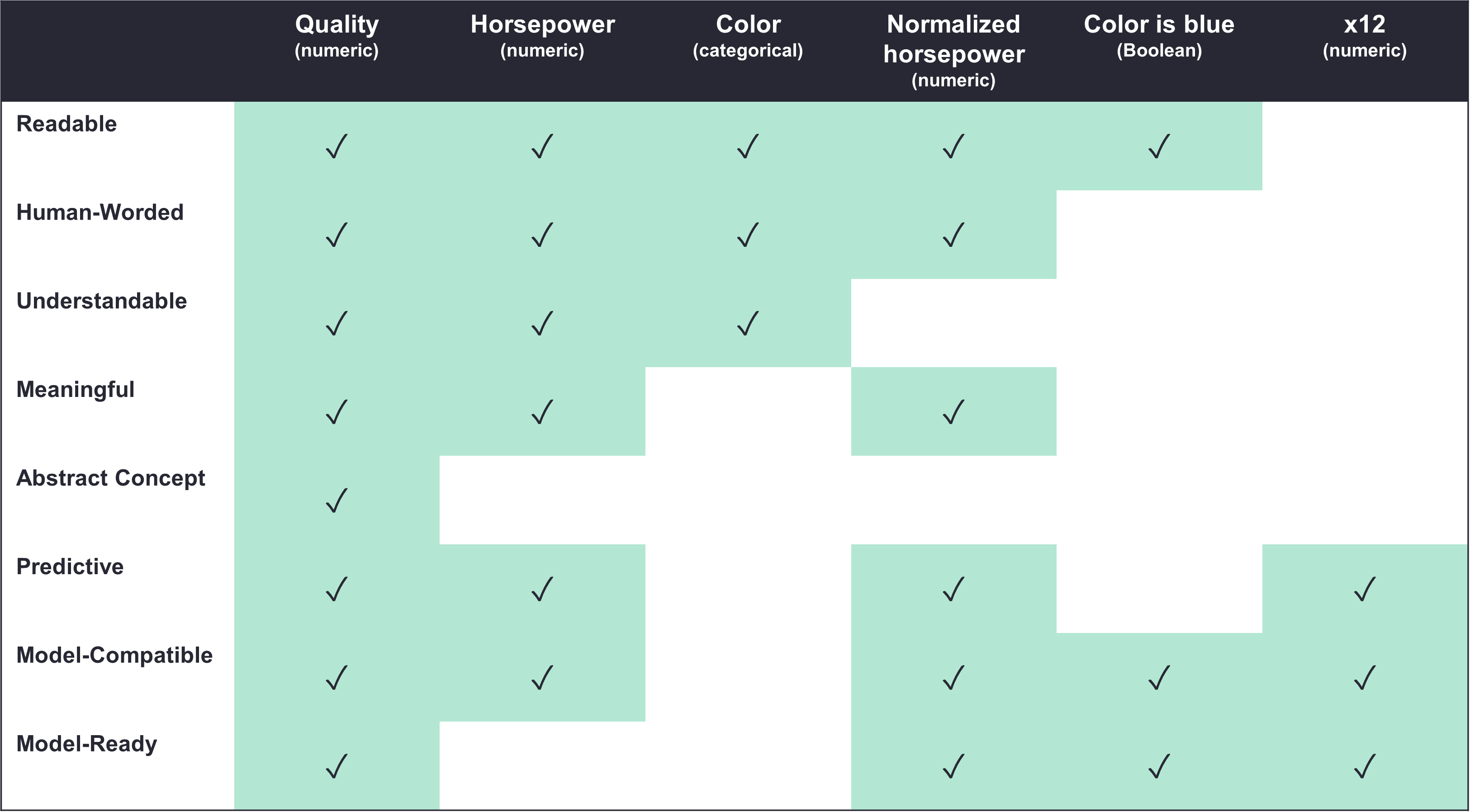}
    \caption{Summary of the feature taxonomy proposed in this paper. For the examples, we use the following hypothetical scenario: a regression model trained on normalized data to predict the maximum speed of the car. \texttt{Quality} is a composite feature computed based on other features, and \texttt{x12} is a arbitrary predictive engineered feature.}
    \label{fig:taxonomy}
\end{figure*}

\subsection{Interpretable Feature Work}
Our methodological review found some existing work on formalizing and quantifying the interpretability of features.

Yadav et. al. \cite{yadav_explanation_2017} surveyed 35 users on Amazon Mechanical Turk to quantify the interpretability of features. Users were provided with a definition of each feature, along with the formula used to calculate its value. Then, users were asked to calculate the value of the feature on unseen data. This metric captures what we in this paper call \property{simulatability}. 

Zhang et. al. \cite{zhang_dissonance_2019} ran a pair of user studies to determine how similar machine-learning-generated features are to those deemed important to humans on image data. They refer to this similarity metric as \textit{dissonance}. Participants were asked to select the super-pixels in images that they believed were most important to identifying the object in the image. Then, they were asked to classify images using the super-pixels considered most important by either humans or ML models (the latter were selected using SHAP \cite{lundberg_unified_2017}). This work found that human-selected features were not necessarily more helpful than model-selected ones to human classification. 

Hong et. al. \cite{hong_human_2020} found through a series of interviews that part of the interpretability workflow includes a \textit{focus on features}. The interviewees in this study stated that investigating features was a useful way to judge the plausibility of a model. They also found that when a model is to be used for decision-making, actionable features may be more useful.

While this existing work motivates the need for interpretable features and offers a few general metrics of feature interpretability, it does not capture the full spectrum of properties that may contribute to the usefulness of features to users. In our work, we offer a more nuanced set of feature properties.

%% file: Figures/literature_review_table.tex
\begin{table*}[ht]
\centering
\caption{Summarized results of the methodological literature review on interpretable features. Papers were counted towards a category if it was explicitly referred to in the text. }
\label{tab:literature_review}
\begin{tabular}{{p{0.6\linewidth}P{0.1\linewidth}p{0.2\linewidth}}}
\hline
\textbf{Comments on Feature Interpretability}                                                                                & \textbf{Number of Papers} & \textbf{Sample Papers} \\ \hline
\textbf{No Elaboration:} Mentions interpretable features without any elaboration or citations on meaning or purpose                   & 47                        &        -{}-{}-               \\
\textbf{Generating IFs:} Proposes an ML architecture or feature engineering/selection approach that results in interpretable features & 49                        &     \cite{cheng_flock_2015} \cite{kim_mind_2015} \cite{shi_self-supervised_2020}                  \\
\textbf{Importance of IFs:} Discusses or claims the general importance of interpretable features                                          & 47                        &   \cite{bhatt_explainable_2020} \cite{choi_ten_2020} \cite{hong_human_2020}                     \\
\textbf{Defines IFs:} Provides some context-specific or general definition or metric for interpretable features.                      & 30                        &  \cite{daniels_scenarionet_2018} \cite{mathew_polar_2020} \cite{nargesian_learning_2017}                   \\
\textbf{Interpreting Features:} Proposes an algorithm for interpreting the meaning of latent neural network features                              & 23                        &  \cite{duan_automated_2018} \cite{guan_towards_2019} \cite{zhang_extracting_2020}                       \\

\textbf{Relies on IFs:} States that the methods in the paper rely on interpretable features to be effective                           & 18                        & \cite{jitkrittum_distinguishing_2016} \cite{saha_language_2019} \cite{sokol_glass-box_2018}                        \\
\textbf{Falls short of IFs:} Explicitly states that the methods in the paper did not achieve interpretable features                   & 5                         &  \cite{beltzung_real-time_2020} \cite{chen_neural_2019} \cite{doron_discovering_2019}                \\
\textbf{Concerns about IFs:} Describes potential drawbacks to interpretable features                                                  & 2                         &   \cite{harder_interpretable_2020} \cite{sheng_anatomy_2018}                     \\\hline

\end{tabular}
\end{table*}

%% file: textv2/taxonomy.tex
\section{Taxonomy of Machine Learning Features} \label{sec:taxonomy}

%\input{Figures/taxonomy_table}

\input{Figures/forest_cover}

\begin{figure*}
    \centering
    \includegraphics[width=0.6\linewidth]{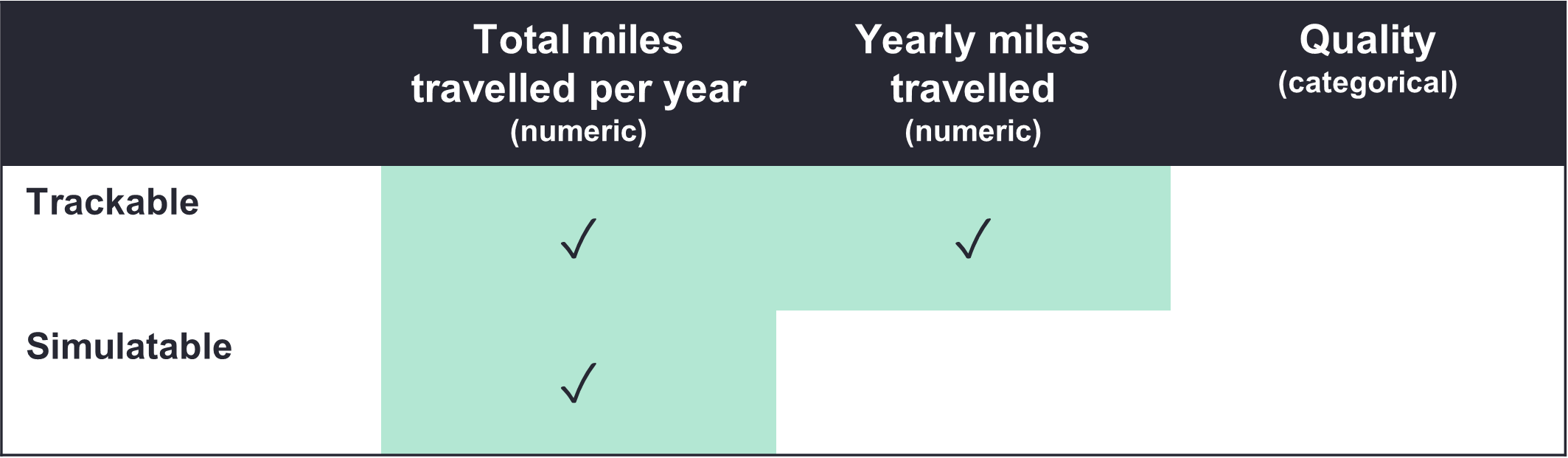}
    \caption{Summary of the feature properties related to tracking feature values back to raw data. \texttt{Quality} is a composite feature computed based on other features, and we assume raw time series data of miles travelled per day over time. \texttt{Yearly miles travelled} is not considered simulatable because it is unclear if this is an average, mean, trend, or another value.}
    \label{fig:taxonomy2}
\end{figure*}

In this section, we formalize and build on the current literature on ML features to develop a thorough and nuanced taxonomy of feature properties. Through this process, we aim to develop guidelines for ML developers to reference when selecting, engineering, and presenting ML features for specific use cases. We motivate our proposed taxonomy through experiences in the real-world domains, and our findings from our formal literature review. Figures \ref{fig:taxonomy} and \ref{fig:taxonomy2} summarize our taxonomy, and provide examples from a hypothetical use case of predicting the maximum speed of a car. 

We do not claim this to be an exhaustive list of feature properties --- we have selected a set that resulted directly from lessons learned while working in our \num domains. We also do not include properties that are specific to the meaning of a feature. For example, we do not include properties such as causal, actionable, inherent \cite{miller_explanation_2018}, or extrinsic features \cite{miller_explanation_2018}. While these kinds of properties are essential to understand when designing explanations, they warrant a separate discussion. 

Different domains and use cases will benefit from different feature properties. Of particular importance is whether the goal is to understand the ML model or the real-world domain. Therefore, we split our discussion of properties into three sections, one for each of these use-cases and one for both. \\

\subsection{Properties Related to the Model-Ready Feature Space}
In this section, we list properties that are most relevant to the model-ready feature space, and are most likely to be necessary when used by individuals most interested in the model (\user{developers}, \user{theorists}, and some \user{ethicists}).
\begin{itemize}
    \item \property{\textbf{Predictive}} features correlate with the prediction target. Features that are not predictive are not expected to improve a model's performance. Feature selection and engineering methodology aim to produce a set of predictive features. 
    
    Features may be predictive due to a directly causal link --- for example, referencing Figure \ref{fig:taxonomy}, it is likely that the horsepower is a predictive feature when predicting the maximum speed of a car. Predictiveness may also result from indirect correlations; for example, we assume car color is not predictive for maximum speed, but it could be if, by coincidence or other reason, faster cars come in similar colors.
    
    \item \property{\textbf{Model-Compatible}} features are in the types or formats that are supported by the model's architecture. Put another way, model-compatible features can be fed into the model to get a prediction without causing errors. For example, many models such a linear regression cannot accept categorical features by default --- these features must be one-hot-encoded or converted to numerical codes. Additionally, many models cannot handle missing data, so imputation is required. 
    
    \item \property{\textbf{Model-Ready}} features are in the final format expected by the model, as used during model training and intended by model developers. This includes both compatibility and performance-based feature engineering. Model-ready features are always \property{model-compatible}. Put another way, model-ready features can both be put into the model without error and will result in the most accurate prediction possible for the model. This will include the required transforms for model-compatibility, as well as transforms such as standardization or normalization. Many of these transformations will reduce the interpretability of features, while improving the performance of the model. 
\end{itemize}

\subsection{Properties Related to the Interpretable Feature Space}
In this section, we list properties that are most relevant to the interpretable feature space. They may be necessary when users are more interested in reality than the model (\user{decision makers}, \user{impacted users}, and \user{ethicists}).
\begin{itemize}
    \item \property{\textbf{Readable}} features are written using a language that allows users to understand what is being referred to, and do not use any codes except those that are readily understood by the users. Concretely, for a feature to be readable it must take from a common, known vocabulary. Across the literature this is generally considered a bare minimum for interpretability. Examples of unreadable features include those using unfamiliar codes  such as \texttt{x12}, unfamiliar acronyms such as \texttt{M\_miss\_DOB} instead of \texttt{Mother is missing date of birth}, or features such as those generated by PCA (unless they are accompanied by an explanation of which input features they came from). Readable features include simple English words and phrases (\texttt{Age}), codes that are readily understood by the target audience (\texttt{cm}), and descriptions of feature engineering operations (\texttt{log(humidity)} \cite{khurana_feature_2018}).
    
    \item \property{\textbf{Human-worded}} features are described in the way most natural for the users. Human-worded features are always \property{readable}. 
    
    We found this property to be essential in the child welfare domain. For example, users were confused by the one-hot encoding language used by the model, so our visualizations displayed them as categorical features (\texttt{gender -> MALE} instead of \texttt{gender is male -> TRUE}). The one-hot encoded versions are \property{readable}, in that they do not use unclear codes, but are still not natural and take additional time to parse. Additionally, child welfare screeners preferred Boolean features presented as negative and positive statements (\texttt{the child has no siblings} rather than \texttt{the child has siblings -> FALSE}).
    
    \item \property{\textbf{Understandable}} features refer to real-world metrics that users can reason about. The extent to which a feature is understandable will depend on the context and users' expertise, but generally this category will not include engineered features that are the results of mathematical operations unless they are commonly referred to concepts. For example, \texttt{age} may be understandable, but \texttt{log(humidity)} \cite{khurana_feature_2018} may not be. 
\end{itemize}

\subsection{Properties Relevant to All Feature Spaces}
These feature properties may be useful both in developing/debugging and using machine learning, and therefore may be desired by all ML users.

\begin{itemize}
    \item \property{\textbf{Meaningful}} features are features that users have reason to believe actually correlate with or are related to the target variable. 
    %Meaningful features are always \property{understandable}. 
    This property may be essential for usable, trustworthy ML that does not confuse end-users. While this property relates closely to \property{predictive}, some features may be predictive as a result of spurious correlations and therefore may not be meaningful to users. Conversely, some features may seem to users like they should correlate with the prediction target, but not actually have any predictive power. ML models that use only meaningful features may generalize better \cite{zhang_dissonance_2019}, so this property may be of interest to ML developers and theorists as well.
    
    As an example, in our case study in child welfare screening, one domain expert said 
    \begin{quote}
        \textit{``... 2 parents have missing date-of-birth is shown as [significantly decreasing risk] which I can't imagine is protective.''}
    \end{quote} 
    In this case, the Boolean feature \texttt{parents missing date of birth} is understandable, but still causes confusion because it is unclear how and why this should correlate with risk.
        
    \item \property{\textbf{Abstract Concepts}} are features that may not be directly measurable, but are calculated through some (usually) domain-expert defined combination of the original dataset features. 
    %These features should be designed to be \property{meaningful}. 
    These features take the form of abstract concepts that use the language of the intended ML users. While they can improve the interpretability of ML models, they may also improve performance. For example, when evaluating massive open online courses (MOOCs) in the domain of education, abstract concepts such as \texttt{participation} and \texttt{achievement} \cite{deboer_changing_2014} are used, which in turn are hand-crafted from metrics such as hours spent watching videos and homework attempts (for the former) and test and homework grades (for the latter).  
    
    \item \property{\textbf{Trackable}} features come with a clear data lineage --- they can be associated accurately with the raw data they were calculated from. For example, in the EHR domain, we found that features like \texttt{mean(heart rate)} may be more useful if presented alongside the original time series heart rate data that the mean was calculated from \cite{cheng_vbridge_2021}. Similarly, in the cybersecurity domain, information about potential DGA attacks were more useful when provided along with the relevant DNS logs \cite{arnaldo_holy_2019}.
    
    Another form of trackability is knowing when a feature value is the result of imputation (for example, the value represents a mean or synthetic value, rather than an actual collected data point), and therefore does not link to any real data. This property was requested by the satellite monitors \cite{liu_mtv_2021}. 
    
    \item \property{\textbf{Simulatable}} features are those where the user has enough information and knowledge to accurately recompute the feature from raw data --- or \textit{simulate} the feature computation process. \property{Simulatable} features are always \property{trackable}.
    
    For example, in the MOOC domain, a feature described as \texttt{test grade over time} is \property{trackable}, as it is clearly computed from raw test grades, but \textbf{not} \property{simulatable}, as it is unclear what the exact formula used to calculate this information is. It could refer to the slope of the trend line of grades, or the difference between the most recent grade and the average grade.
\end{itemize}

%\subsubsection{Other Feature Descriptors}
%\note{maybe remove this section for space}
%Here, we list other feature properties described by the literature which are factors of the features themselves. It may be useful to consider whether features have these properties when deciding how to provide explanations.
%\begin{enumerate}
    
 %   \item \inherent{} features, as defined by Prasada \?  are properties determined by the kind of thing an instance is. For example, for a instance of type dog, \texttt{number of legs = 4} would be an inherent feature, as having four legs is a property of (most) dogs \?. 
    
  %  \item \extrinsic{} features, as defined by Prasada \? are those that are not \inherent{} to the type of a thing --- for example, a dog may have a feature \texttt{wearing a collar = True}, which is not a direct consequence of it being a dog, but rather of a human putting a collar on it. As found by Prasada, explanations \note{elaborate on importance of these two terms to explanation}
    
   % \item \property{Actionable} features \? are those that the user can act on the change. 

%\end{enumerate}

\section{Feature Transforms} \label{sec:transforms}
Having introduced a set of feature properties that may be desirable, we will now approach the topic of how to achieve these properties when they are necessary.

Some properties must be considered during the feature selection phase. For example,  \property{predictiveness} and \property{meaningfulness} are attributes of the feature meaning itself, and can only be provided by carefully selecting such features. \property{Understandability} will also in part come from careful feature selection.

However, many other features can be offered through feature engineering, transforms, and other processing steps on pre-selected features. \property{Readability}, \property{trackability}, and \property{simulatability} can often be offered through hand-writing feature definitions or providing relevant visualizations. \property{Model-compatibility} and \property{model-readiness} are offered through traditional feature engineering. \property{Human-wordedness} and to some extent \property{understandability} can also be offered by feature engineering, though possibly in the reverse direction.

In this section, we consider two categories of ML feature transforms. \textit{Model-ready transforms}, or those that take features from the original feature space to the model-ready feature space, have received extensive coverage in the literature \cite{khalid_survey_2014} \cite{zheng_feature_2018}. In this section, we suggest a subset of possible \textit{interpretable transforms}, or those that take features from the original feature space to the interpretable features space. These transforms are often not covered in feature engineering literature and libraries. For example, \texttt{sklearn} \cite{pedregosa_scikit-learn_2011} has a \texttt{OneHotEncoder} transform that one-hot encodes categorical data; the \texttt{pandas} \cite{the_pandas_development_team_pandas-devpandas_2020} \texttt{get\_dummies()} function does the same. However, neither of these libraries have built in functionality to ``undo'' this encoding, computing a categorical feature from a set of Boolean features. 
%Similarly, while many feature engineering libraries can easily offer imputation capabilities (such as \texttt{sklearn}'s \texttt{SimpleImputer}), flags indicating which data have been imputed must be coded in and tracked manually \cite{brownlee_add_2020}. 

Table \ref{tab:forest-cover} provides examples of these transforms from the forest covertype dataset \cite{dua_uci_2017}. This dataset includes 12 input features and one target variable. Each observation in this dataset refers to a 30 by 30 meter patch in Roosevelt National Forest, and includes information such as elevation, sunlight levels, and soil type. The intended target variable is the type of forest cover in that area.

\subsection{Model-ready transforms} Traditionally, most work on feature engineering has focused on the model-ready feature state. These transforms generally have the purpose of 1) converting the data to a form that is compatible with the model, such as one-hot encoding categorical data or imputing missing data; 2) improving performance by factoring in an understanding of the model's methodology, such as normalizing features for models that are sensitive to scale or using dimensionality reduction to avoid overfitting; or 3) improving performance by factoring in an understanding of the domain. 

Model-ready transforms will often reduce the interpretability of features. For example, in the child welfare domain, we learned that one-hot encoding reduces the \property{human-wordedness} of features, and binning numeric features may reduce the \property{understandability} of features if the bins are not semantically meaningful \cite{zytek_sibyl_2021}. Additionally, in our experience with satellite signal monitoring, we found that imputing data can reduce \property{trackability} if users do not know which data is real. 

\subsection{Interpretable transforms}
Some model-ready transforms may naturally lead to interpretable features --- especially those that factor in domain knowledge to improve model performance --- but other times, feature interpretability is improved either by undoing model-ready transforms, or introducing new transforms for use in explanations.

Here, we introduce transforms that may improve feature interpretability, along with real-life examples from our experiences. 

\begin{itemize}
\item \textbf{\transform{Converting to Categorical}.} One-hot encoded features are not \property{human-worded}. For example, \texttt{gender is female --- TRUE} should be replaced with \texttt{gender --- FEMALE}. This transform becomes especially important as the cardinality of the categorical feature grows.

\item \textbf{\transform{Semantic Binning}.} Binning numeric features based on meaningful, real-world distinctions can improve both the interpretability and performance of models. For example, in the child welfare domain, users frequently referenced and relied on the age categories, which were binned based on child development (infant, toddler, teenager, etc.).   

This kind of binning is often more interpretable than binning based on statistical metrics (such as ensuring equal bin widths or sizes), as is often done to improve model performance. In the forest covertype example, consider the difference between binning elevation based on ecologically significant elevation zones such as \texttt{foothills} vs \texttt{alpine}, compared to generating bins of equal widths, resulting in bins such as \texttt{2525m - 3192m}.

\item \textbf{\transform{Flagged Imputations}.} Explanations that display imputed (synthetic) data can confuse users. In our experience with satellite monitoring, the raw data signals included missing data that was then imputed. When our visualizations did not distinguish between real data and imputed data, users were confused. 

\item \textbf{\transform{Aggregated Numeric Features}.} Numeric features may be more interpretable in an aggregated format, especially when the model takes several very closely related metrics. In the forest covertype example, we combine the horizontal and vertical distance metrics to get total distance to hydrology. In the child welfare example, the overload of data that screeners experienced could be remedied in part by summing together measures, such as physical, sexual, and emotional abuse referrals, into a total referral count metric.

\item \textbf{\transform{Categorical Feature Hierarchies}.} Categorical features are often involved in hierarchical relationships with other features. Both model performance and model interpretability can be improved by selecting the right granularity for a given domain. In our forest covertype example, all soil types present in the dataset belong to exactly one of eight geologic soil zones. Therefore, we could either present the individual soil type, or the soil zone, depending on which is more useful to users. 

\item \textbf{\transform{Combining to Abstract Concepts}.} An extension of the previous two transforms: abstract concepts carefully crafted from data by domain experts can be much faster and easier to parse than large numbers of features. This is generally done by combining a set of features with a hand-crafted formula. In our forest covertype example, an abstract, human-intuitive feature may be \texttt{light level}, computed by aggregating the three \texttt{hillshade} features that quantify the illumination of a point at different times of day.

\item \textbf{\transform{Reversing Scaling}.} Standardized or normalized features are almost never interpretable. Features should  be displayed in their unstandardized format in explanations, unless the relative position of a value is meaningful.

\item \textbf{\transform{Reversing Feature Engineering}.} Some transforms done to improve model performance may need to be undone to improve interpretability. For example, a feature represented by \texttt{sqrt(value)} may need to be squared to get \texttt{value}.

\item \textbf{\transform{Connecting to Raw Data}.} Sometimes, the raw data (for example, the original signal data) is more meaningful to users than the engineered features. For example, in the EHR domain, the patient pulse signal itself is often more useful than the feature \texttt{MEAN(pulse)}. Therefore, explanations may be more useful if they display the signal itself, linked to the model features.

\end{itemize}

\section{Discussion}

\subsection{Using Feature Properties}

We will now revisit a few of the sample domains introduced in section \ref{sec:introduction}, to give complete examples of how the proposed taxonomy in this paper could be used to improve ML usability.

In the domain of child welfare, we identified --- through interviews and field observations --- that our target users were \user{decision makers} who were domain experts in child welfare, but were not expected to have any prior understanding of data science or ML. Their primary objective is to make screening decisions, and want explanations of ML predictions to aid with this task. They trust their existing decision making process, and mainly want ML to highlight information they may have missed. For these reasons, we might say that child welfare screening requires features that are \property{meaningful}, \property{understandable}, and \property{human-worded}. We can achieve these properties in part through  \transform{feature selection} and \transform{semantic binning}, while ensuring features we present are \transform{categorical} and \transform{unstandardized}. 

In the domain of healthcare and making predictions on electronic health records, we identified our users as \user{decision makers} looking to improve outcomes during surgery. Again, the users were domain experts without ML knowledge. They were used to making decisions with full access to patient data, in particular time series for metrics such as vital signs. Therefore, features needed to be \property{simulatable}, which is provided by \transform{connecting the features to data} and including \transform{impute flags}.

\FloatBarrier

Similarly to these two examples, we encourage practitioners introducing ML to new domains to collaborate with end-users to understand which feature properties will be necessary to optimize usability.

\subsection{Next Steps}
In this paper, we motivated the need for a more nuanced consideration of feature interpretability, and propose a first version of a formal taxonomy of feature properties. This taxonomy should be iterated on based on the experiences of other researchers, to capture the full spectrum of properties that may be useful. 

Additionally, a more formal connection between the needs of users and the useful feature properties should be developed. For example, \property{meaningful} features were useful to child welfare screeners because they already trusted and did not intend to greatly modify their existing decision making process. In other domains, where users are more interested in learning about the domain and developing new decision making workflows, they may rely less on \property{meaningful} features and instead be content with \property{understandable} features. 

In addition, as described in section \ref{sec:transforms}, future work will develop systems that make converting to the interpretable feature space as seamless a process as converting to the model-ready feature space.

\subsection{Risks}

While interpretable features may be necessary to make a model useful, there are always risks involved in transforming to the interpretable feature space. Providing explanations using anything other than the model-ready feature space risks lowering the fidelity of the explanation. Some of the transformations can also greatly (and possibly intentionally) bias the explanations. Selecting formulas to generate abstract concepts and binning numerical data are such ways to intentionally hide certain patterns being used by the model. For example, consider a crime recidivism prediction ML model that will increase the risk of recidivism for an individual if they are black. A developer could maliciously include the race feature in a broad, abstract concept called \texttt{socioeconomic factors} that hides the effects of race itself. 

Additionally, forcing features to have interpretability properties may reduce model performance. In particular, limiting a feature set to only those features seen as meaningful to humans will prevent the model from discovering potentially unknown or unexpected patterns in the data. Some of these patterns may relate to real-world correlations or causal paths that would have been very valuable to the scientific knowledge of the field. It is essential for practitioners to understand the trade-offs between different feature spaces, and make context-aware decisions when selecting features.

Finally, as noted by two papers from our literature review \cite{harder_interpretable_2020} \cite{sheng_anatomy_2018}, using interpretable features may reduce data privacy in cases where the features describe sensitive information. In such cases, it may be necessary to avoid interpretable properties such as \property{readability} in the interest of preserving privacy.

%% file: Figures/forest_cover.tex
% \usepackage{booktabs}
% \usepackage{multirow}
\begin{sidewaystable*}[htb]
\caption{Example of feature transforms using the Forest Cover dataset}
\label{tab:forest-cover}
\footnotesize
\begin{tabular}{>{\raggedright}p{0.12\linewidth}>{\raggedright}p{0.15\linewidth}>{\raggedright}p{0.2\linewidth}>{\raggedright}p{0.08\linewidth}>{\raggedright}p{0.08\linewidth}p{0.08\linewidth}>{\raggedright\arraybackslash}p{0.08\linewidth}}
\hline
\textbf{Feature Transform}        & \textbf{Feature Name}            & \textbf{Feature Description}                                                                                    & \textbf{Feature Type} & \multicolumn{3}{l}{\textbf{Sample Values}}                               \\ \hline
Original         & Elevation                        & Elevation in meters                                                                                             & Numeric               & 3179                    & 3123                 & 2157                    \\
                                  & Horizontal Distance To Hydrology & Horizontal distance to nearest surface water features                                                               & Numeric               & 450                     & 218                  & 85                      \\
                                  & Vertical Distance To Hydrology   & Vertical distance to nearest surface water features                                                                 & Numeric               & 56                      & 21                   & 10                      \\
                                  & Hillshade 9am                    & Hillshade index at 9am, summer solstice                                                                         & Numeric               & 156                     & 228                  & 210                     \\
                                  & Hillshade Noon                   & Hillshade index at noon, summer soltice                                                                         & Numeric               & 231                     & 229                  & 131                     \\
                                  & Hillshade 3pm                    & Hillshade index at 3pm, summer solstice                                                                         & Numeric               & 211                     & 187                  & 103                     \\
                                  & Wilderness area                  & Wilderness area designation                                                                                     & Categorical           & Comache Peak            & Comache Peak         & Rawah                   \\ \hline
\multicolumn{7}{l}{\textbf{Model-Ready Transforms}} \\                                                                                                                                                                                                                                      \\ \hline
Statistical Binning               & Elevation Range                  & Uniform-width bins for Elevation                                                                                & Categorical           & Medium (2525m-3192m)    & Medium (2525m-3192m) & Low (1859m-2525m)       \\
One-Hot Encoding & Area Rawah            & Wilderness area is Rawah                                                                                        & Boolean               & FALSE                   & FALSE                & TRUE                    \\
                                  & Area Neota            & Wilderness area is Neota                                                                                        & Boolean               & FALSE                   & FALSE                & FALSE                   \\
                                  & Area Comache Peak     & Wilderness area is Comache Peak                                                                                 & Boolean               & TRUE                    & TRUE                 & FALSE                   \\
                                  & Area Cache la Poudre  & Wilderness area is Cache la Poudre                                                                              & Boolean               & FALSE                   & FALSE                & FALSE                   \\
Standardization                   & Elevation Standardized           & Standardized Elevation                                                                                          & Numeric               & 0.784                   & 0.584                & -2.87                   \\
PCA              & PCA 1                            & Feature 1 from PCA                                                                                              & Numeric               & -1.965                  & 2.278                & -0.851                  \\
                                  & PCA 2                            & Feature 2 from PCA                                                                                              & Numeric               & 2.265                   & 1.547                & 3.095                   \\ \hline
\multicolumn{7}{l}{\textbf{Interpretable Transforms}}  \\                                                                                                                                                                                                                                 \\ \hline
Semantic Binning                  & Elevation Zone                   & Colorado Life Zone corresponding to elevation                                                                   & Categorical           & Subalpine               & Subalpine            & Foothills               \\
Flagged Imputed                   & Elevation Flag                   & Flags if elevation value was imputed                                                                            & Boolean               & FALSE                   & FALSE                & FALSE                   \\
Aggregated Numerical              & Distance from Hydrology          & Total distance from hydrology, computed from horizontal and vertical distances & Numeric               & 453                     & 219                  & 85                      \\
Hierarchical Categorical           & Soil Geologic Zone               & Geologic zone of soil type (8 total)                                                                            & Categorical           & igneous and metamorphic & glacial              & igneous and metamorphic \\
Abstract Concepts                 & Light Level                      & Overall light level (Sum of hillshade features)                                                                & Categorical           & High                    & High                 & Medium                  \\ \hline
\end{tabular}
\end{sidewaystable*}

%% file: textv2/conclusion.tex
\section{Conclusion}

In this paper, we motivated the need for a more nuanced discussion on what we refer to as the interpretable feature space. We proposed a partial, formal taxonomy for interpretable feature properties, backed by our real-world experience in \num{} domains. We also suggested an initial list of feature transforms that can provide these properties. We encourage the community to continue to consider and extend the ways in which features need to be interpretable for different contexts, especially those that rely on decision makers without ML experience.

%% file: main.bbl
\begin{thebibliography}{10}

\bibitem{arnaldo_holy_2019}
I.~Arnaldo and K.~Veeramachaneni.
\newblock The {Holy} {Grail} of "{Systems} for {Machine} {Learning}": {Teaming}
  humans and machine learning for detecting cyber threats.
\newblock {\em ACM SIGKDD Explorations Newsletter}, 21(2):39--47, Nov. 2019.

\bibitem{beltzung_real-time_2020}
L.~Beltzung, A.~Lindley, O.~Dinica, N.~Hermann, and R.~Lindner.
\newblock Real-{Time} {Detection} of {Fake}-{Shops} through {Machine}
  {Learning}.
\newblock In {\em 2020 {IEEE} {International} {Conference} on {Big} {Data}
  ({Big} {Data})}, pages 2254--2263, Dec. 2020.

\bibitem{bhatt_explainable_2020}
U.~Bhatt, A.~Xiang, S.~Sharma, A.~Weller, A.~Taly, Y.~Jia, J.~Ghosh, R.~Puri,
  J.~M.~F. Moura, and P.~Eckersley.
\newblock Explainable machine learning in deployment.
\newblock In {\em Proceedings of the 2020 {Conference} on {Fairness},
  {Accountability}, and {Transparency}}, {FAT}* '20, pages 648--657, New York,
  NY, USA, Jan. 2020. Association for Computing Machinery.

\bibitem{chen_neural_2019}
X.~Chen, Q.~Lin, C.~Luo, X.~Li, H.~Zhang, Y.~Xu, Y.~Dang, K.~Sui, X.~Zhang,
  B.~Qiao, W.~Zhang, W.~Wu, M.~Chintalapati, and D.~Zhang.
\newblock Neural {Feature} {Search}: {A} {Neural} {Architecture} for
  {Automated} {Feature} {Engineering}.
\newblock In {\em 2019 {IEEE} {International} {Conference} on {Data} {Mining}
  ({ICDM})}, pages 71--80, Nov. 2019.
\newblock ISSN: 2374-8486.

\bibitem{cheng_vbridge_2021}
F.~Cheng, D.~Liu, F.~Du, Y.~Lin, A.~Zytek, H.~Li, H.~Qu, and K.~Veeramachaneni.
\newblock {VBridge}: {Connecting} the {Dots} {Between} {Features} and {Data} to
  {Explain} {Healthcare} {Models}.
\newblock {\em IEEE Transactions on Visualization and Computer Graphics}, pages
  1--1, 2021.
\newblock arXiv: 2108.02550.

\bibitem{cheng_flock_2015}
J.~Cheng and M.~S. Bernstein.
\newblock Flock: {Hybrid} {Crowd}-{Machine} {Learning} {Classifiers}.
\newblock In {\em Proceedings of the 18th {ACM} {Conference} on {Computer}
  {Supported} {Cooperative} {Work} \& {Social} {Computing}}, {CSCW} '15, pages
  600--611, New York, NY, USA, Feb. 2015. Association for Computing Machinery.

\bibitem{choi_ten_2020}
M.~Choi, L.~M. Aiello, K.~Z. Varga, and D.~Quercia.
\newblock Ten {Social} {Dimensions} of {Conversations} and {Relationships}.
\newblock In {\em Proceedings of {The} {Web} {Conference} 2020}, {WWW} '20,
  pages 1514--1525, New York, NY, USA, Apr. 2020. Association for Computing
  Machinery.

\bibitem{daniels_scenarionet_2018}
Z.~A. Daniels and D.~Metaxas.
\newblock {ScenarioNet}: {An} {Interpretable} {Data}-{Driven} {Model} for
  {Scene} {Understanding}.
\newblock {\em IJCAI Workshop on Explainable Artificial Intelligence (XAI)
  2018}, July 2018.

\bibitem{deboer_changing_2014}
J.~DeBoer, A.~D. Ho, G.~S. Stump, and L.~Breslow.
\newblock Changing “{Course}”: {Reconceptualizing} {Educational}
  {Variables} for {Massive} {Open} {Online} {Courses}.
\newblock {\em Educational Researcher}, 43(2):74--84, Mar. 2014.
\newblock Publisher: American Educational Research Association.

\bibitem{doron_discovering_2019}
M.~Doron, I.~Segev, and D.~Shahaf.
\newblock Discovering {Unexpected} {Local} {Nonlinear} {Interactions} in
  {Scientific} {Black}-box {Models}.
\newblock In {\em Proceedings of the 25th {ACM} {SIGKDD} {International}
  {Conference} on {Knowledge} {Discovery} \& {Data} {Mining}}, {KDD} '19, pages
  425--435, New York, NY, USA, July 2019. Association for Computing Machinery.

\bibitem{doshi-velez_towards_2017}
F.~Doshi-Velez and B.~Kim.
\newblock Towards {A} {Rigorous} {Science} of {Interpretable} {Machine}
  {Learning}.
\newblock {\em arXiv:1702.08608 [cs, stat]}, Mar. 2017.
\newblock arXiv: 1702.08608.

\bibitem{dua_uci_2017}
D.~Dua and C.~Graff.
\newblock {UCI} machine learning repository, 2017.
\newblock Publisher: University of California, Irvine, School of Information
  and Computer Sciences.

\bibitem{duan_automated_2018}
J.~Duan, Z.~Zeng, A.~Oprea, and S.~Vasudevan.
\newblock Automated {Generation} and {Selection} of {Interpretable} {Features}
  for {Enterprise} {Security}.
\newblock In {\em 2018 {IEEE} {International} {Conference} on {Big} {Data}
  ({Big} {Data})}, pages 1258--1265, Dec. 2018.

\bibitem{guan_towards_2019}
C.~Guan, X.~Wang, Q.~Zhang, R.~Chen, D.~He, and X.~Xie.
\newblock Towards a {Deep} and {Unified} {Understanding} of {Deep} {Neural}
  {Models} in {NLP}.
\newblock In {\em International {Conference} on {Machine} {Learning}}, pages
  2454--2463. PMLR, May 2019.
\newblock ISSN: 2640-3498.

\bibitem{harder_interpretable_2020}
F.~Harder, M.~Bauer, and M.~Park.
\newblock Interpretable and {Differentially} {Private} {Predictions}.
\newblock {\em Proceedings of the AAAI Conference on Artificial Intelligence},
  34(04):4083--4090, Apr. 2020.
\newblock Number: 04.

\bibitem{hong_human_2020}
S.~R. Hong, J.~Hullman, and E.~Bertini.
\newblock Human {Factors} in {Model} {Interpretability}: {Industry}
  {Practices}, {Challenges}, and {Needs}.
\newblock {\em Proceedings of the ACM on Human-Computer Interaction},
  4(CSCW1):1--26, May 2020.

\bibitem{jitkrittum_distinguishing_2016}
W.~Jitkrittum, Z.~Szabo, K.~Chwialkowski, and A.~Gretton.
\newblock Distinguishing distributions with interpretable features, June 2016.
\newblock Conference Name: International Conference on Machine Learning (ICML):
  Data-Efficient Machine Learning workshop Meeting Name: International
  Conference on Machine Learning (ICML): Data-Efficient Machine Learning
  workshop Place: New York, USA Publisher: International Conference on Machine
  Learning (ICML): Data-Efficient Machine Learning workshop Volume: 2016.

\bibitem{khalid_survey_2014}
S.~Khalid, T.~Khalil, and S.~Nasreen.
\newblock A survey of feature selection and feature extraction techniques in
  machine learning.
\newblock In {\em 2014 {Science} and {Information} {Conference}}, pages
  372--378, Aug. 2014.

\bibitem{khurana_feature_2018}
U.~Khurana, H.~Samulowitz, and D.~Turaga.
\newblock Feature {Engineering} for {Predictive} {Modeling} {Using}
  {Reinforcement} {Learning}.
\newblock {\em Proceedings of the AAAI Conference on Artificial Intelligence},
  32(1), Apr. 2018.
\newblock Number: 1.

\bibitem{kim_mind_2015}
B.~Kim, J.~A. Shah, and F.~Doshi-Velez.
\newblock Mind the {Gap}: {A} {Generative} {Approach} to {Interpretable}
  {Feature} {Selection} and {Extraction}.
\newblock {\em NIPS}, 2015.
\newblock Accepted: 2017-05-26T15:24:08Z Publisher: Neural Information
  Processing Systems Foundation Inc.

\bibitem{liu_mtv_2021}
D.~Liu, S.~Alnegheimish, A.~Zytek, and K.~Veeramachaneni.
\newblock {MTV}: {Visual} {Analytics} for {Detecting}, {Investigating}, and
  {Annotating} {Anomalies} in {Multivariate} {Time} {Series}.
\newblock {\em arXiv:2112.05734 [cs]}, Dec. 2021.
\newblock arXiv: 2112.05734.

\bibitem{lundberg_unified_2017}
S.~M. Lundberg and S.-I. Lee.
\newblock A {Unified} {Approach} to {Interpreting} {Model} {Predictions}.
\newblock In {\em Advances in {Neural} {Information} {Processing} {Systems}},
  volume~31, page~10, Long Beach, California, 2017. Curran Associates Inc.

\bibitem{mathew_polar_2020}
B.~Mathew, S.~Sikdar, F.~Lemmerich, and M.~Strohmaier.
\newblock The {POLAR} {Framework}: {Polar} {Opposites} {Enable}
  {Interpretability} of {Pre}-{Trained} {Word} {Embeddings}.
\newblock In {\em Proceedings of {The} {Web} {Conference} 2020}, {WWW} '20,
  pages 1548--1558, New York, NY, USA, Apr. 2020. Association for Computing
  Machinery.

\bibitem{miller_explanation_2018}
T.~Miller.
\newblock Explanation in {Artificial} {Intelligence}: {Insights} from the
  {Social} {Sciences}.
\newblock {\em arXiv:1706.07269 [cs]}, Aug. 2018.
\newblock arXiv: 1706.07269.

\bibitem{molnar_interpretable_2020}
C.~Molnar, G.~Casalicchio, and B.~Bischl.
\newblock Interpretable {Machine} {Learning} -- {A} {Brief} {History},
  {State}-of-the-{Art} and {Challenges}.
\newblock {\em arXiv:2010.09337 [cs, stat]}, Oct. 2020.
\newblock arXiv: 2010.09337.

\bibitem{nargesian_learning_2017}
F.~Nargesian, H.~Samulowitz, U.~Khurana, E.~B. Khalil, and D.~Turaga.
\newblock Learning {Feature} {Engineering} for {Classification}.
\newblock In {\em Proceedings of the {Twenty}-{Sixth} {International} {Joint}
  {Conference} on {Artificial} {Intelligence}}, pages 2529--2535, Melbourne,
  Australia, Aug. 2017. International Joint Conferences on Artificial
  Intelligence Organization.

\bibitem{the_pandas_development_team_pandas-devpandas_2020}
T.~pandas~development team.
\newblock pandas-dev/pandas: {Pandas}, Feb. 2020.

\bibitem{pedregosa_scikit-learn_2011}
F.~Pedregosa, G.~Varoquaux, A.~Gramfort, V.~Michel, B.~Thirion, O.~Grisel,
  M.~Blondel, P.~Prettenhofer, R.~Weiss, V.~Dubourg, J.~Vanderplas, A.~Passos,
  and D.~Cournapeau.
\newblock Scikit-learn: {Machine} {Learning} in {Python}.
\newblock {\em Journal of Machine Learning Research}, 12:2825--2830, 2011.

\bibitem{plohmann_comprehensive_2016}
D.~Plohmann, K.~Yakdan, M.~Klatt, E.~Gerhards-Padilla, and J.~Bader.
\newblock A {Comprehensive} {Measurement} {Study} of {Domain} {Generating}
  {Malware}.
\newblock In {\em 25th {USENIX} {Security} {Symposium}}, volume~25, page~17,
  Austin, TX, 2016.

\bibitem{preece_stakeholders_2018}
A.~Preece, D.~Harborne, D.~Braines, R.~Tomsett, and S.~Chakraborty.
\newblock Stakeholders in {Explainable} {AI}.
\newblock In {\em {AAAI} {FSS}-18: {Artificial} {Intelligence} in {Government}
  and {Public} {Sector}}, page~6, Arlington, Virginia, 2018.

\bibitem{saha_language_2019}
K.~Saha, S.~C. Kim, M.~D. Reddy, A.~J. Carter, E.~Sharma, O.~L. Haimson, and
  M.~De~Choudhury.
\newblock The {Language} of {LGBTQ}+ {Minority} {Stress} {Experiences} on
  {Social} {Media}.
\newblock {\em Proceedings of the ACM on Human-Computer Interaction},
  3(CSCW):89:1--89:22, Nov. 2019.

\bibitem{sheng_anatomy_2018}
Y.~Sheng, S.~Tata, J.~B. Wendt, J.~Xie, Q.~Zhao, and M.~Najork.
\newblock Anatomy of a {Privacy}-{Safe} {Large}-{Scale} {Information}
  {Extraction} {System} {Over} {Email}.
\newblock In {\em Proceedings of the 24th {ACM} {SIGKDD} {International}
  {Conference} on {Knowledge} {Discovery} \& {Data} {Mining}}, {KDD} '18, pages
  734--743, New York, NY, USA, July 2018. Association for Computing Machinery.

\bibitem{shi_self-supervised_2020}
W.~Shi, G.~Huang, S.~Song, Z.~Wang, T.~Lin, and C.~Wu.
\newblock Self-{Supervised} {Discovering} of {Interpretable} {Features} for
  {Reinforcement} {Learning}.
\newblock {\em IEEE Transactions on Pattern Analysis and Machine Intelligence},
  pages 1--1, 2020.
\newblock Conference Name: IEEE Transactions on Pattern Analysis and Machine
  Intelligence.

\bibitem{sokol_glass-box_2018}
K.~Sokol and P.~Flach.
\newblock Glass-{Box}: {Explaining} {AI} {Decisions} {With} {Counterfactual}
  {Statements} {Through} {Conversation} {With} a {Voice}-enabled {Virtual}
  {Assistant}.
\newblock In {\em Proceedings of the {Twenty}-{Seventh} {International} {Joint}
  {Conference} on {Artificial} {Intelligence}}, pages 5868--5870, Stockholm,
  Sweden, July 2018. International Joint Conferences on Artificial Intelligence
  Organization.

\bibitem{veeramachaneni_towards_2014}
K.~Veeramachaneni, U.-M. O'Reilly, and C.~Taylor.
\newblock Towards {Feature} {Engineering} at {Scale} for {Data} from {Massive}
  {Open} {Online} {Courses}.
\newblock {\em arXiv:1407.5238 [cs]}, July 2014.
\newblock arXiv: 1407.5238.

\bibitem{yadav_explanation_2017}
A.~Yadav, A.~Rahmattalabi, E.~Kamar, P.~Vayanos, M.~Tambe, and V.~L. Noronha.
\newblock Explanation {Systems} for {Influence} {Maximization} {Algorithms}.
\newblock In {\em 3rd {International} {Workshop} on {Social} {Influence}
  {Analysis} ({SocInf} 2017)}, page~12, 2017.

\bibitem{zhang_extracting_2020}
Q.~Zhang, X.~Wang, R.~Cao, Y.~N. Wu, F.~Shi, and S.-C. Zhu.
\newblock Extracting an {Explanatory} {Graph} to {Interpret} a {CNN}.
\newblock {\em IEEE Transactions on Pattern Analysis and Machine Intelligence},
  pages 1--1, 2020.
\newblock Conference Name: IEEE Transactions on Pattern Analysis and Machine
  Intelligence.

\bibitem{zhang_dissonance_2019}
Z.~Zhang, J.~Singh, U.~Gadiraju, and A.~Anand.
\newblock Dissonance {Between} {Human} and {Machine} {Understanding}.
\newblock {\em Proceedings of the ACM on Human-Computer Interaction},
  3(CSCW):1--23, Nov. 2019.

\bibitem{zheng_feature_2018}
A.~Zheng and A.~Casari.
\newblock {\em Feature {Engineering} for {Machine} {Learning}: {Principles} and
  {Techniques} for {Data} {Scientists}}.
\newblock "O'Reilly Media, Inc.", Mar. 2018.
\newblock Google-Books-ID: sthSDwAAQBAJ.

\bibitem{zytek_sibyl_2021}
A.~Zytek, D.~Liu, R.~Vaithianathan, and K.~Veeramachaneni.
\newblock Sibyl: {Understanding} and {Addressing} the {Usability} {Challenges}
  of {Machine} {Learning} {In} {High}-{Stakes} {Decision} {Making}.
\newblock {\em IEEE Transactions on Visualization and Computer Graphics}, pages
  1--1, 2021.
\newblock Conference Name: IEEE Transactions on Visualization and Computer
  Graphics.

\end{thebibliography}
